\documentclass[letterpaper]{article} % DO NOT CHANGE THIS
\PassOptionsToPackage{table}{xcolor}
\usepackage[preprint]{aaai2027}  % Show authors and suppress the review-only notice. 
\usepackage[hyphens]{url}  % DO NOT CHANGE THIS
\usepackage{graphicx} % DO NOT CHANGE THIS
\usepackage{natbib}  % DO NOT CHANGE THIS AND DO NOT ADD ANY OPTIONS TO IT
\usepackage{caption} % DO NOT CHANGE THIS AND DO NOT ADD ANY OPTIONS TO IT
\usepackage{amsmath}
\usepackage{amssymb}
\usepackage{algorithm}
\usepackage{algorithmic}
\usepackage{tikz}
\usepackage{booktabs}
\usepackage{multirow}
\usepackage{makecell}
\usepackage{array}
\usepackage{tabularx}
\usepackage{adjustbox}

\definecolor{GroupGray}{RGB}{235,235,235}
\definecolor{OursYellow}{RGB}{255,248,210}

\usepackage{pifont}
\newcommand{\cmark}{\ding{51}}
\newcommand{\respair}[2]{#1 / #2}
\newcommand{\best}[1]{\textbf{#1}}
\newcommand{\second}[1]{\underline{#1}}

\providecommand{\miss}{}
\renewcommand{\miss}{\textbf{\textemdash}}

\providecommand{\respairmissright}[1]{}
\renewcommand{\respairmissright}[1]{(#1,\,\miss)}

\providecommand{\respairmissleft}[1]{}
\renewcommand{\respairmissleft}[1]{(\miss,\,#1)}

\definecolor{VADRowGray}{RGB}{232,232,232}
\definecolor{OursRowYellow}{RGB}{255,242,204}
\definecolor{AblOneYellow}{RGB}{255,247,224}
\definecolor{AblTwoYellow}{RGB}{255,250,236}
\definecolor{AblThreeYellow}{RGB}{255,253,245}
\definecolor{DropOrange}{RGB}{210,135,65}

\newcolumntype{L}{>{\raggedright\arraybackslash}p{2.45cm}}
\newcolumntype{C}{>{\centering\arraybackslash}p{0.72cm}}
\newcolumntype{Y}{>{\centering\arraybackslash}X}
\newcolumntype{P}{>{\centering\arraybackslash}p{0.54cm}}

\title{FreqAnchorAD: Language-Free Zero-Shot Anomaly Detection via Frequency-Deviation Anchoring}

\author{
Jianfeng Qiu\textsuperscript{\rm 1},
Peiyuan Li\textsuperscript{\rm 1},
Juan Xie\textsuperscript{\rm 1},
Xueliang Ma\textsuperscript{\rm 1},\\
Sihang Zhou\textsuperscript{\rm 2},
Yanning Hou\textsuperscript{\rm 2*},
Ke Xu\textsuperscript{\rm 1*}
}

\affiliations{
\textsuperscript{\rm 1}Anhui University, Hefei, China\\
\textsuperscript{\rm 2}National University of Defense Technology, Changsha, China
}

\begin{document}

\maketitle

\begingroup
\renewcommand{\thefootnote}{\fnsymbol{footnote}}
\footnotetext[1]{Corresponding authors.}
\endgroup
\begin{abstract}
Zero-shot anomaly detection (ZSAD) aims to detect anomalous samples and localize defective regions in unseen target domains without using target training data. 
Many recent ZSAD methods build on pretrained vision models, particularly CLIP, and construct normal and anomaly references using textual prompts or learnable visual representations. Text-guided and visual-reference methods perform anomaly discrimination primarily in spatial feature spaces, where subtle changes in texture, boundaries, and local structures are difficult to distinguish from normal appearance variations.
Although such defects are inconspicuous in the spatial domain, they can disrupt local texture regularity or boundary continuity, thereby inducing distinguishable response deviations across different frequency bands.
However, these frequency-dependent characteristics are not explicitly modeled by existing ZSAD methods.
Our image-domain analysis reveals that local defects exhibit spatial-frequency deviations from their normal references across low-, middle-, and high-frequency bands, indicating that anomaly evidence is not universally dominated by high-frequency responses. Motivated by this observation, we propose \textbf{FreqAnchorAD}, a frequency-aware framework centered on organizing frequency-enhanced responses for anchor-relative anomaly discrimination.
Specifically, we propose a Local Frequency Compensation Module (LFCM) to enhance intermediate patch tokens with local spatial-frequency cues.
Moreover, we introduce the Frequency-Deviation Anchor Projector (FDAP), the core discrimination module that organizes the enhanced responses along a source-derived channel coordinate and measures anomaly evidence through relative similarity to normal and anomaly anchors.
Finally, we design Asymmetric Anchor Supervision (AAS) to stabilize normal-anchor alignment while preserving diverse anomaly patterns.
Experiments on thirteen industrial and medical benchmarks show that \textbf{FreqAnchorAD} achieves state-of-the-art mean performance on both image-level anomaly recognition and pixel-level defect localization.

\end{abstract}

\section{Introduction}

Anomaly detection~\cite{AD1,AD2,AD3,AD4} aims to detect abnormal samples and localize defective regions by identifying deviations from normal patterns.
Conventional methods~\cite{PaDiM,CS-Flow,CutPaste} usually rely on target category normal samples to learn specific normality, but collecting sufficient and clean normal data for each target category is often challenging in real-world deployments.
This motivates zero-shot anomaly detection (ZSAD), which aims to detect and localize anomalies without using any target-domain training samples~\cite{musc,SAA}.

\begin{figure*}[t]
    \centering
    \includegraphics[
        width=\textwidth,
        height=0.46\textheight,
        keepaspectratio
    ]{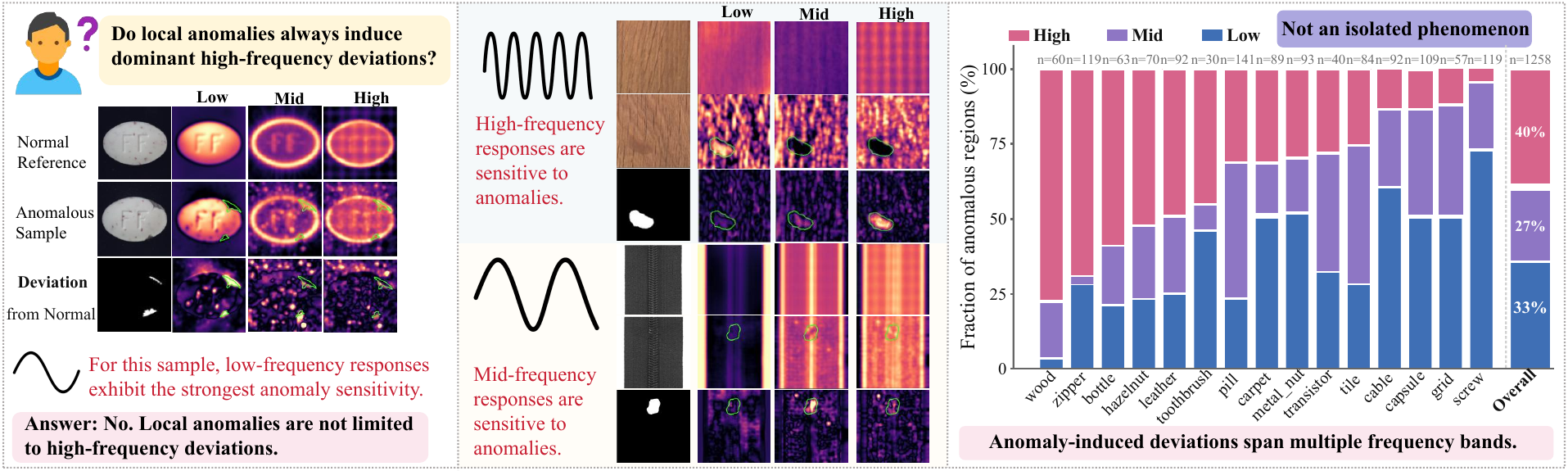}
    \caption{
    Motivation for frequency-aware anomaly deviation modeling.
    Spatial-frequency analysis relative to category-specific normal references shows that low-, mid-, and high-frequency deviations can each dominate anomaly localization for different images and categories.
    These observations indicate that local anomalies are better characterized as frequency-dependent deviations from normal references than as universally high-frequency responses, motivating frequency-aware representation and its integration with reference-based anomaly discrimination.
    }
    \label{fig:intro_frequency_deviation}
\end{figure*}

This problem is challenging due to domain shifts across unseen categories, diverse anomaly appearances, and subtle local defects.
Most ZSAD methods build on CLIP~\cite{CLIP}, whose transferable representations provide useful priors for unseen categories. 
Text-guided methods~\cite{WinCLIP,APRIL-GAN,CLIP-AD,VCP,FiLo,AdaCLIP,AA-CLIP} construct normal and anomaly references through textual prompts. 
In contrast, VisualAD~\cite{VisualAD} removes the language branch and learns purely visual references, resulting in a simpler framework with strong ZSAD performance. 
Despite their different reference construction strategies, both paradigms discriminate anomalies primarily in spatial feature spaces. 
In these spaces, subtle texture, boundary, and local structural changes can be difficult to distinguish from normal appearance variations. 
Frequency-domain modeling can complement spatial representations by explicitly characterizing the spectral deviations induced by these changes. 
FE-CLIP~\cite{FE-CLIP} introduces frequency cues into CLIP-based anomaly detection to enhance the representation of subtle local defects. 
However, its frequency cues mainly serve to enhance visual features, rather than constructing a channel-spectral space for directly comparing patch representations with normal and anomalous token references, limiting the explicit exploitation of subtle spectral deviations for anomaly discrimination.
To address this gap, we propose FreqAnchorAD, a frequency-aware framework for ZSAD that organizes frequency-enhanced responses in a structured anchor space for reference-based anomaly discrimination.

To characterize local anomaly deviations, we conduct a post hoc analysis of spatial frequency in the image domain, as shown in Fig.~\ref{fig:intro_frequency_deviation}. Using same-category normal training images only as diagnostic references, we construct normal-calibrated low-, mid-, and high-frequency deviation maps and identify the dominant band according to its sensitivity to anomalous regions, quantified by pixel-level AUROC. High-frequency deviations are most frequent ($39.75\%$), while low- and mid-frequency deviations jointly account for $60.25\%$, motivating collaborative multi-band modeling rather than reliance on a single frequency band.

Motivated by the observed multi-band deviations, we propose FreqAnchorAD, a frequency-aware ZSAD framework that combines local spatial frequency enhancement with source-derived channel organization for anomaly discrimination relative to normal and anomaly anchors.
Specifically, the Local Frequency Compensation Module (LFCM) injects local spatial-frequency cues into patch representations, while the Frequency-Deviation Anchor Projector (FDAP), as the core discrimination module, organizes the enhanced channel responses along a channel coordinate
constructed from source-domain normal and anomalous responses and applies a spectral transform for anchor-relative discrimination.
The resulting channel-spectral components characterize variations along the source-derived channel coordinate rather than spatial frequency bands of the input image or token grid.
Asymmetric Anchor Supervision (AAS) further regularizes the resulting channel-spectral anchor space by stabilizing normal alignment while preserving diverse anomaly patterns.

The main contributions of this paper are summarized as follows:

\begin{itemize}

\item We provide an image-domain spatial-frequency analysis showing that low-, mid-, and high-frequency bands exhibit complementary anomaly-localization sensitivity across images and categories, motivating collaborative multi-band modeling of visual representations.

\item We introduce source-derived channel canonicalization, which constructs a consistent channel coordinate from the response statistics of normal and anomalous source images for channel-spectral modeling.

\item We propose FreqAnchorAD, an FDAP-centered frequency-aware ZSAD framework that integrates local spatial-frequency compensation, source-derived channel-spectral anchor projection, and asymmetric anchor supervision for robust relative normal/anomaly discrimination.

\item Extensive experiments on thirteen industrial and medical benchmarks demonstrate that FreqAnchorAD achieves competitive image-level anomaly recognition and strong pixel-level defect localization under the ZSAD setting.

\end{itemize}

\section{Related Work}
\label{sec:related_work}

\subsection{Reference-based Zero-shot Anomaly Detection}
\label{sec:related_clip_zsad}

Reference-based ZSAD methods aim to detect anomalies by comparing test samples with normal or abnormal references in a pretrained representation space. 
WinCLIP~\cite{WinCLIP} constructs normal/anomaly references from hand-crafted textual prompts, while AnomalyCLIP~\cite{AnomalyCLIP} introduces learnable anomaly semantics to refine textual anomaly references. 
VisualAD~\cite{VisualAD} further revisits the necessity of the CLIP language branch and constructs visual reference features for language-free anomaly discrimination.
However, they mainly operate on CLIP-derived spatial domain features, making it difficult to explicitly characterize subtle signal-level perturbations such as texture irregularities, boundary changes, and micro-structural defects. 
Our work constructs a source-derived channel coordinate and performs anchor-relative anomaly discrimination in the resulting channel-spectral space.

\subsection{Frequency-aware Visual Representation Learning}
\label{sec:related_frequency_ad}

Frequency-domain modeling has been widely explored to complement spatial visual representations~\cite{DFD,FAIR,FDNM,Wave-MambaAD,fdp,wavelet}.
FFC~\cite{FFC} captures non-local and cross-scale interactions through Fourier units. 
FcaNet~\cite{FcaNet} uses multi-spectral DCT components for channel attention, while GFNet~\cite{GFNet} models long-range dependencies through global frequency-domain filtering. 
FE-CLIP~\cite{FE-CLIP} extends frequency modeling to ZSAD by incorporating frequency-aware feature extraction and local frequency statistics into the CLIP visual encoder. 
However, it mainly uses frequency information for feature enhancement, without explicitly organizing anomaly-induced frequency deviations relative to normal and anomaly references. 
In contrast, FreqAnchorAD combines local spatial-frequency compensation with anchor-relative channel-spectral modeling.

\section{Method}
\label{sec:method}

\begin{figure*}[t]
\centering
\includegraphics[width=\textwidth]{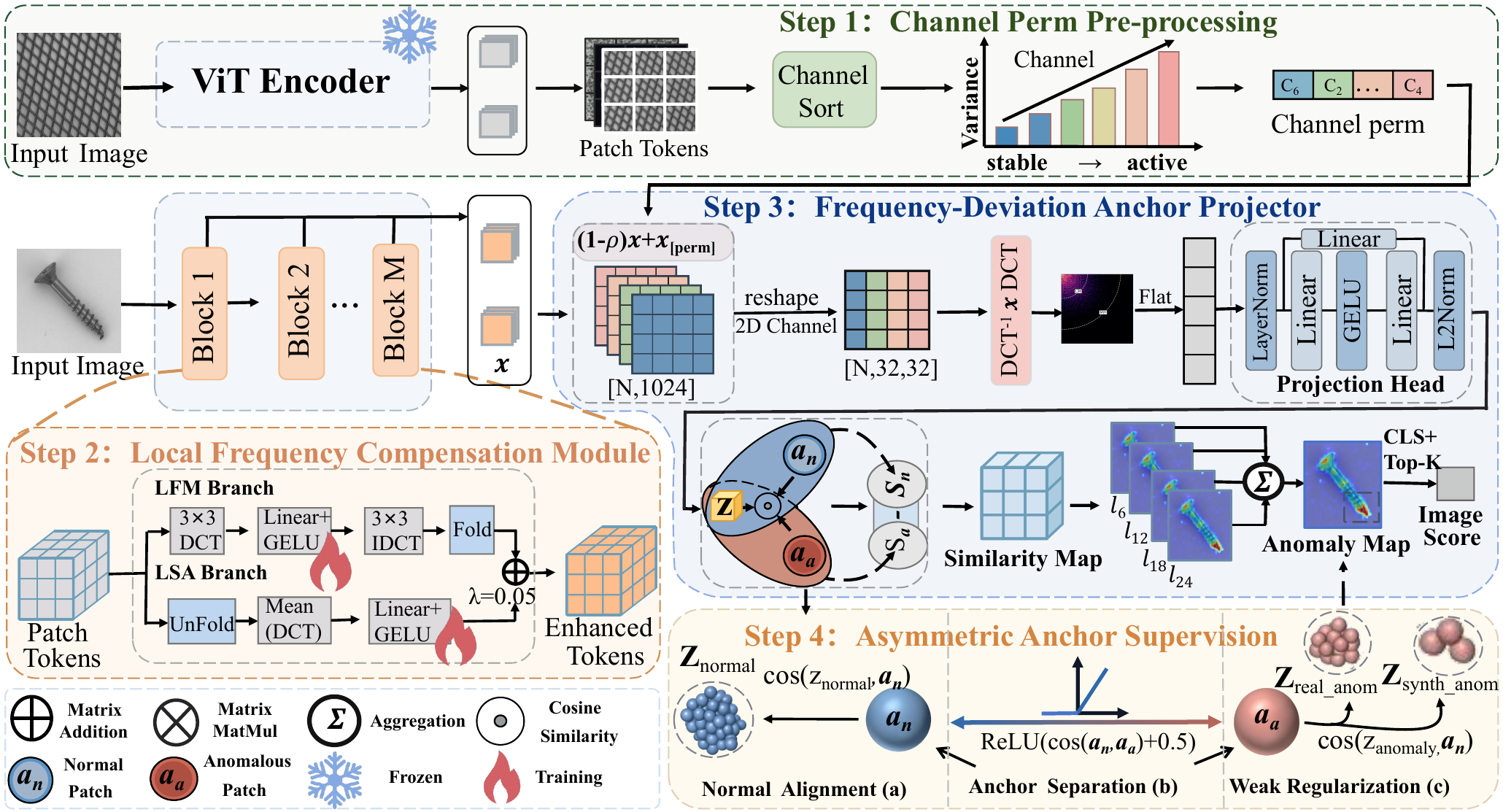}
\caption{
Overview of FreqAnchorAD.
A frozen vision encoder extracts multi-layer patch tokens, LFCM enhances local
spatial-frequency cues, and FDAP projects them into a channel-spectral anchor
space for relative normal/anomaly scoring.
AAS regularizes this space during training, while aggregated layer-wise maps
produce the final anomaly map and image score.
}
\label{fig:frame}
\end{figure*}

\subsection{Problem Setting and Overview}
\label{sec:method_overview}

We follow the zero-shot anomaly detection (ZSAD) setting, where the model is trained on auxiliary source categories and directly evaluated on unseen target categories. 
Let $\mathcal{C}_{s}$ and $\mathcal{C}_{t}$ denote the source and target category sets, with $\mathcal{C}_{s}\cap\mathcal{C}_{t}=\varnothing$. 
During evaluation, no target-domain training images are used.

Motivated by these frequency-dependent deviations, we propose FreqAnchorAD, a language-free, frequency-aware framework centered on FDAP-based anchor-relative discrimination. 
LFCM provides defect-sensitive local spatial-frequency cues, 
FDAP organizes the enhanced responses in a source-derived channel-spectral anchor space, and AAS stabilizes the resulting normal/anomaly anchor discrimination.

Given an image $\mathbf{I}$, the frozen backbone extracts intermediate patch tokens $\mathbf{X}^{l}\in\mathbb{R}^{B\times N\times D}$ from layers $\mathcal{L}$, where $N=H\times W$. 
LFCM enriches patch tokens with local frequency cues, FDAP maps the enhanced tokens into a source-derived channel-spectral anchor space for anchor-relative anomaly scoring, and AAS regularizes this space by stabilizing normal alignment while preserving diverse anomaly-side deviations.
The whole pipeline operates only on visual encoder features, without textual prompts, cross-modal alignment, target-domain statistics, or test-time adaptation.
The overview of the proposed model is illustrated in Fig.~\ref{fig:frame}.

\subsection{Local Frequency Compensation Module}
\label{sec:lfcm}

The frozen vision encoder provides transferable visual representations, but its intermediate patch tokens may not explicitly preserve local spatial-frequency deviations on the token grid.
To address this limitation, we introduce the Local Frequency Compensation Module (LFCM) at each selected layer to enhance the sensitivity of visual features to texture, boundary, and structural perturbations.
Given patch tokens $\mathbf{X}^{\ell}\in\mathbb{R}^{B\times N\times D}$, we first reshape them into a spatial token grid $\mathbf{X}^{\ell}_{g}\in\mathbb{R}^{B\times D\times H\times W}$, where $N=H\times W$. 
LFCM contains two complementary branches: Local Frequency Modulation (LFM) uses DCT-based modulation on non-overlapping local windows to capture compact spectral perturbations, while Local Spectral Aggregation (LSA) summarizes DCT statistics from sliding neighborhoods to capture regional structural deviations.
The two branches produce frequency responses $\mathbf{F}^{\ell}_{\mathrm{LFM}}$ and $\mathbf{F}^{\ell}_{\mathrm{LSA}}$, which are fused with the original token grid by
\begin{equation}
\widetilde{\mathbf{X}}^{\ell}_{g}
=
(1-\lambda)\mathbf{X}^{\ell}_{g}
+
\lambda
\left(
\mathbf{F}^{\ell}_{\mathrm{LFM}}
+
\mathbf{F}^{\ell}_{\mathrm{LSA}}
\right).
\end{equation}
The enhanced grid is then flattened back to patch tokens $\widetilde{\mathbf{X}}^{\ell}\in\mathbb{R}^{B\times N\times D}$.
This residual compensation preserves the pretrained representation while injecting local spatial-frequency evidence for anchor-based anomaly scoring.

\subsection{Frequency-Deviation Anchor Projector}
\label{sec:fdap}

LFCM enhances patch tokens with frequency-sensitive local responses. FDAP further organizes these responses along a source-derived channel coordinate for normal/anomaly anchor discrimination.
To this end, FDAP maps the enhanced tokens into a channel-spectral anchor space and quantifies anomaly evidence through relative anchor similarity.
However, the channels of pretrained visual features have no inherent spatial or frequency ordering.
Consequently, directly applying a channel-spectral transform would produce spectral coordinates determined by arbitrary channel indices rather than by a stable response structure.
We address this issue by constructing a source-derived channel canonicalization from source-domain normal and anomalous responses.

For each selected layer $\ell$, we collect all patch tokens from the
selected source split:
\begin{equation}
\mathbf{X}^{\ell}_{s}
=
\operatorname{Concat}
\left(
\mathbf{X}^{\ell}_{s,n},
\mathbf{X}^{\ell}_{s,a}
\right),
\end{equation}
where $\mathbf{X}^{\ell}_{s,a}$ denotes all patch tokens extracted from
anomalous source images rather than only those inside anomalous regions.
The channel-wise variance and permutation are computed as
\begin{equation}
\begin{aligned}
v^{\ell}_{c}
&=
\frac{1}{M_{\ell}}
\sum_{i=1}^{M_{\ell}}
\left(
X^{\ell}_{s,ic}
\right)^2
-
\left(
\frac{1}{M_{\ell}}
\sum_{i=1}^{M_{\ell}}
X^{\ell}_{s,ic}
\right)^2,\\
\boldsymbol{\pi}^{\ell}
&=
\operatorname{Argsort}_{\mathrm{asc}}
\left(
\mathbf{v}^{\ell}
\right).
\end{aligned}
\end{equation}

where $M_\ell$ denotes the number of source patch tokens in
$\mathbf{X}_s^\ell$.
The ascending permutation orders channels from relatively stable low-variance
responses to more active high-variance responses, thereby establishing a
reproducible source-derived channel coordinate before the channel-spectral
transform.

Given an LFCM-enhanced patch token $\widetilde{\mathbf{x}}^{\ell}_{i}$, its variance-ordered representation is defined as
\begin{equation}
\mathbf{x}^{\ell,\pi}_{i}
=
\widetilde{\mathbf{x}}^{\ell}_{i}
\left[
\boldsymbol{\pi}^{\ell}
\right].
\end{equation}
The objective of FDAP is to represent target tokens in a unified source-derived channel coordinate system for anomaly scoring.
Accordingly, the inference rule of FDAP is defined as full channel canonicalization, corresponding to $\rho=1$, where all target samples are mapped into the same deterministic canonical channel space.

During training, directly enforcing the fully canonicalized representation may introduce an abrupt shift from the pretrained feature space to the reordered channel space. To alleviate this optimization difficulty, we introduce soft channel canonicalization:

\begin{equation}
\overline{\mathbf{x}}^{\ell}_{i}
=
(1-\rho)
\widetilde{\mathbf{x}}^{\ell}_{i}
+
\rho
\mathbf{x}^{\ell,\pi}_{i},
\end{equation}

where $\rho<1$ controls the degree of canonicalization only during training. This strategy preserves part of the pretrained channel structure while gradually adapting the projection head and anchors to the canonical channel space. 
Unless otherwise specified, we set the training coefficient to $\rho=0.6$ throughout all experiments.

The canonicalized channel sequence is arranged into an
$H_c\times W_c$ computational grid, where $H_cW_c=D$.
To determine the grid shape consistently across different channel
dimensions, we adopt a dimension-balanced factorization rule:
\begin{equation}
(H_c,W_c)
=
\underset{\substack{hw=D\\h\geq w}}
{\operatorname{arg\,min}}
\,|h-w|.
\label{eq:balanced_channel_layout}
\end{equation}
This rule limits the cross-row stride of the ordered sequence while balancing the spectral resolution along the two transform axes.
For $D=1024$, this rule yields a balanced $32\times32$ DCT grid based on the source-derived channel order rather than any intrinsic spatial structure; when no nontrivial factorization exists, the layout reduces to $D\times1$.

The organized channel representation is then transformed using a two-dimensional DCT~\cite{DCT}:
\begin{equation}
\mathbf{D}^{\ell}_{i}
=
\operatorname{DCT}_{2D}
\left(
\operatorname{Reshape}_{H_c\times W_c}
\left(
\overline{\mathbf{x}}^{\ell}_{i}
\right)
\right).
\end{equation}
The DCT is applied over the source-derived channel coordinate rather than the spatial layout of the input image, producing a channel-spectral representation of different response-variation rates.

We then flatten the resulting channel-spectral representation and apply layer normalization:
\begin{equation}
\mathbf{u}^{\ell}_{i}
=
\operatorname{LN}
\left(
\operatorname{Flatten}
\left(
\mathbf{D}^{\ell}_{i}
\right)
\right).
\end{equation}
The normalized channel-spectral representation is projected into the channel-spectral anchor space through an MLP branch and a residual projection branch:
\begin{equation}
\mathbf{z}^{\ell}_{i}
=
\operatorname{Norm}_{2}
\left(
\mathcal{F}_{\ell}
\left(
\mathbf{u}^{\ell}_{i}
\right)
+
\mathcal{P}_{\ell}
\left(
\mathbf{u}^{\ell}_{i}
\right)
\right),
\end{equation}
where $\mathcal{F}_{\ell}$ denotes a layer-specific two-layer MLP projection head, and $\mathcal{P}_{\ell}$ denotes the residual projection.

Given the normal anchor $\mathbf{a}^{\ell}_{n}$ and anomaly anchor $\mathbf{a}^{\ell}_{a}$, the anomaly score of the $i$-th patch is computed as
\begin{equation}
s^{\ell}_{i}
=
\cos
\left(
\mathbf{z}^{\ell}_{i},
\mathbf{a}^{\ell}_{a}
\right)
-
\cos
\left(
\mathbf{z}^{\ell}_{i},
\mathbf{a}^{\ell}_{n}
\right).
\end{equation}
Thus, FDAP measures anomaly evidence through anchor-relative deviations in the channel-spectral space rather than through absolute responses in a predefined spectral band.
The patch scores are finally rearranged according to their spatial positions to obtain the layer-wise anomaly map $\mathbf{S}^{\ell}$.

\subsection{Training Objective}
\label{sec:training_objective}
We introduce Asymmetric Anchor Supervision (AAS) to stabilize the
channel-spectral anchor space learned by FDAP.
Normal representations are strongly aligned with the normal anchor, while
real and synthetic anomaly-side representations are weakly guided toward
the anomaly anchor:
\begingroup
\allowdisplaybreaks[2]
\begin{align}
\mathcal{L}^{\ell}_{n}
&=
1-\frac{1}{|\Omega^{\ell}_{n}|}
\sum_{i\in\Omega^{\ell}_{n}}
\cos(\mathbf{z}^{\ell}_{i},\mathbf{a}^{\ell}_{n}),\\
\mathcal{R}^{\ell}_{a,r}
&=
1-\frac{1}{|\Omega^{\ell}_{a,r}|}
\sum_{i\in\Omega^{\ell}_{a,r}}
\cos(\mathbf{z}^{\ell}_{i},\mathbf{a}^{\ell}_{a}),\\
\mathcal{R}^{\ell}_{a,s}
&=
1-\frac{1}{|\Omega^{\ell}_{a,s}|}
\sum_{i\in\Omega^{\ell}_{a,s}}
\cos(\widetilde{\mathbf{z}}^{\ell}_{i},\mathbf{a}^{\ell}_{a}),
\end{align}
\endgroup
where $\Omega_n^\ell$ and $\Omega_{a,r}^\ell$ index normal and anomaly-labeled
source patches, respectively, while $\Omega_{a,s}^\ell$ indexes synthetic
representations $\widetilde{\mathbf{z}}_i^\ell$ generated by adding Gaussian
noise $\boldsymbol{\epsilon}\sim\mathcal{N}(\mathbf{0},\sigma^2\mathbf{I})$ to
normal frequency-domain features before projection.
% where $\Omega^{\ell}_{a,r}$ contains all patches from anomalous images, and
% $\widetilde{\mathbf{z}}^{\ell}_{i}$ is obtained by adding Gaussian noise
% $\boldsymbol{\epsilon}\sim\mathcal{N}(\mathbf{0},\sigma^2\mathbf{I})$ to a
% normal feature before projection.

The anomaly-side regularization and anchor-separation term are
\begin{align}
\mathcal{R}^{\ell}_{a}
&=
\eta_r\mathcal{R}^{\ell}_{a,r}
+\eta_s\mathcal{R}^{\ell}_{a,s},\\
\mathcal{L}^{\ell}_{\mathrm{sep}}
&=
\max\!\left(
0,
\cos(\mathbf{a}^{\ell}_{n},\mathbf{a}^{\ell}_{a})-m
\right).
\end{align}
The AAS objective is
\begin{equation}
\mathcal{L}_{\mathrm{AAS}}
=
\frac{1}{|\mathcal{L}|}
\sum_{\ell\in\mathcal{L}}
\left(
\mathcal{L}^{\ell}_{n}
+\beta\mathcal{L}^{\ell}_{\mathrm{sep}}
+\gamma\mathcal{R}^{\ell}_{a}
\right),
\end{equation}
where the smaller anomaly-side weight $\gamma$ preserves heterogeneous
anomaly responses.

The layer-wise anomaly maps and image-level score are computed as
\begin{equation}
\begin{aligned}
\mathbf{M}
&=
\sum_{\ell\in\mathcal{L}}\operatorname{Up}(\mathbf{S}^{\ell}),
\qquad
s_{\mathrm{map}}
=
\operatorname{TopKMean}(\mathbf{M},r),
\\
s_{\mathrm{img}}
&=
\lambda_{\mathrm{map}}s_{\mathrm{map}}
+
(1-\lambda_{\mathrm{map}})s_{\mathrm{CLS}}.
\end{aligned}
\end{equation}
For CLIP-based industrial experiments, $s_{\mathrm{CLS}}$ denotes the CLS-anchor score and is fused directly with $s_{\mathrm{map}}$ using $\lambda_{\mathrm{map}}=0.5$. For DINOv2 and all medical configurations, the CLS-anchor head is disabled and $s_{\mathrm{img}}=s_{\mathrm{map}}$.
The overall objective is
\begin{equation}
\mathcal{L}_{\mathrm{total}}
=
\alpha\mathcal{L}_{\mathrm{cls}}
+\mathcal{L}_{\mathrm{seg}}
+\lambda_{\mathrm{AAS}}\mathcal{L}_{\mathrm{AAS}},
\end{equation}
where $\mathcal{L}_{\mathrm{cls}}$ is the image-level BCE loss and
$\mathcal{L}_{\mathrm{seg}}$ combines focal and Dice losses~\cite{focal-loss,dice-loss}.

During training, the pretrained vision encoder is frozen, while LFCM, the FDAP projection heads, and the learnable normal and anomaly anchors are optimized under the AAS objective.
During inference, all learned modules and source-derived channel permutations are fixed. FDAP always performs full canonicalization ($\rho=1$) and does not use text encoders, target-domain statistics, or test-time adaptation.

\section{Experiments}
\label{sec:experiments}

\subsection{Experimental Setup}
\label{sec:exp_setup}

\textbf{Datasets.}
We evaluate FreqAnchorAD on six industrial benchmarks: MVTec AD~\cite{MVTec}, VisA~\cite{visa}, BTAD~\cite{BTAD}, KSDD2~\cite{KSDD2}, DAGM~\cite{DAGM}, and DTD-Synthetic~\cite{DTD-Synthetic}.
We further use seven medical benchmarks: OCT17~\cite{OCT17}, BrainMRI~\cite{BrainMRI}, Brain\_AD~\cite{Brain_AD01,Brain_AD02,Brain_AD03}, HIS~\cite{HIS}, CVC-ClinicDB~\cite{CVC-ClinicDB}, Endo~\cite{Endo}, and Kvasir~\cite{Kvasir}.
Following the cross-dataset ZSAD protocol, VisA is used as the source dataset for all non-VisA targets, while MVTec AD is used for VisA; no target-dataset samples are used during training.

\textbf{Baselines.}
We compare with WinCLIP~\cite{WinCLIP}, APRIL-GAN~\cite{APRIL-GAN}, CLIP-AD~\cite{CLIP-AD}, AnomalyCLIP~\cite{AnomalyCLIP}, AdaCLIP~\cite{AdaCLIP}, FE-CLIP~\cite{FE-CLIP}, and VisualAD~\cite{VisualAD}. FE-CLIP is included only in the industrial comparison due to its limited medical dataset coverage and inconsistent pixel-level metrics.

\textbf{Evaluation metrics.}
We report AUROC, F1-max, and AP for image-level evaluation, and AUROC, F1-max, AP, and AUPRO for pixel-level evaluation.
Industrial tables report all metrics for each benchmark.
Image- and pixel-level metrics are prefixed by I- and P-, respectively.

\textbf{Implementation details.}
We instantiate FreqAnchorAD with frozen CLIP ViT-L/14@336px~\cite{CLIP} and DINOv2 ViT-L/14~\cite{DINO,DINOv2} backbones and resize all inputs to $518 \times 518$.
LFCM is inserted into layers $\mathcal{L}=\{6,12,18,24\}$, and FDAP operates on their patch tokens.
FDAP interpolates between the original and source-derived channel orders with $\rho=0.6$ during training and uses the fully canonicalized order with $\rho=1.0$ during inference.
AAS supervises the corresponding channel spectral anchor spaces.
Additional implementation details are provided in the Appendix.

\subsection{Performance Comparison with SOTA Methods}
\label{sec:main_results}

Tables~\ref{tab:industrial_compact_results} and~\ref{tab:medical_compact_results} report the results on industrial and medical benchmarks, respectively.
FreqAnchorAD achieves state-of-the-art overall performance on industrial benchmarks and consistently improves over the corresponding VisualAD baselines on medical benchmarks.
The stronger pixel-level results indicate that the proposed frequency-aware framework effectively captures subtle local anomalies.
Its consistent gains with both CLIP and DINOv2 further demonstrate good backbone and cross-domain generality.

% ===== Table colors =====
\definecolor{TableGroupGray}{RGB}{255,255,255}
\definecolor{TableOursGray}{RGB}{215,215,215}

% ===== Table macros =====
\providecommand{\cmark}{}
\renewcommand{\cmark}{\textcolor{black}{\ding{51}}}

\providecommand{\xmark}{}
\renewcommand{\xmark}{\textcolor{black}{\ding{55}}}

\providecommand{\best}[1]{\textbf{#1}}
\providecommand{\second}[1]{\underline{#1}}

% Mean results: best in bold, second-best underlined.
\providecommand{\bestmean}[1]{}
\renewcommand{\bestmean}[1]{\textbf{#1}}

\providecommand{\secondmean}[1]{}
\renewcommand{\secondmean}[1]{\underline{#1}}

% Dataset entry: (AUROC, AP)
\providecommand{\respair}[2]{}
\renewcommand{\respair}[2]{(#1,\,#2)}

% Industrial image-level entry: (I-AUROC, I-F1-max, I-AP)
\providecommand{\resimg}[3]{}
\renewcommand{\resimg}[3]{(#1,#2,#3)}

% Industrial pixel-level entry: (P-AUROC, P-F1-max, P-AP, P-AUPRO)
\providecommand{\respix}[4]{}
\renewcommand{\respix}[4]{(#1,#2,#3,#4)}

% Missing values used only by the industrial table.
\providecommand{\IndResMiss}{}
\renewcommand{\IndResMiss}{\textemdash}

% Industrial Avg. entries use table-specific names to avoid preamble conflicts.
\providecommand{\IndAvgImg}[3]{}
\renewcommand{\IndAvgImg}[3]{(#1,\,#2,\,#3)}

\providecommand{\IndAvgPix}[4]{}
\renewcommand{\IndAvgPix}[4]{(#1,\,#2,\,#3,\,#4)}

% Medical Avg. entries.
\providecommand{\AvgOneImg}[3]{}
\renewcommand{\AvgOneImg}[3]{(#1,\,#2,\,#3)}

\providecommand{\AvgOnePix}[4]{}
\renewcommand{\AvgOnePix}[4]{(#1,\,#2,\,#3,\,#4)}

% Metric labels.
\providecommand{\metric}[1]{}
\renewcommand{\metric}[1]{%
  \textit{\fontsize{4.45pt}{5.0pt}\selectfont (#1)}%
}

\providecommand{\avgmetric}[1]{}
\renewcommand{\avgmetric}[1]{%
  \textit{\fontsize{4.05pt}{4.6pt}\selectfont (#1)}%
}

% Move section labels slightly left without changing table alignment.
\providecommand{\sectionshift}[1]{}
\renewcommand{\sectionshift}[1]{\hspace{-0.5cm}\textit{\textbf{#1}}}

% ===== Two compact tables in one float =====
\begin{table*}[t]
\centering
\begingroup

% ===================== Table 1: Industrial =====================
% \label{tab:industrial_compact_results}

\fontsize{5.35pt}{6.55pt}\selectfont
\setlength{\tabcolsep}{0.45pt}
\renewcommand{\arraystretch}{1.12}
\setlength{\arrayrulewidth}{0.32pt}
\arrayrulecolor{black}
\setlength{\minrowclearance}{0pt}
\setlength{\extrarowheight}{0pt}
\begin{tabularx}{\textwidth}{@{}
>{\centering\arraybackslash}p{0.20cm}
>{\raggedright\arraybackslash}p{1.62cm}
*{6}{>{\hsize=1.30\hsize\centering\arraybackslash}X}
| *{4}{>{\hsize=0.55\hsize\centering\arraybackslash}X}
@{}}
\toprule
&
\textbf{Method}
& \textbf{MVTec}
& \textbf{VisA}
& \textbf{BTAD}
& \textbf{KSDD2}
& \textbf{DAGM}
& \textbf{DTD}
& \multicolumn{4}{|c@{}}{\textbf{Avg.}} \\
\midrule
\rowcolor{TableGroupGray}
\multicolumn{2}{@{}c}{\sectionshift{Image-level}}
& \metric{I-AUROC,I-F1,I-AP}
& \metric{I-AUROC,I-F1,I-AP}
& \metric{I-AUROC,I-F1,I-AP}
& \metric{I-AUROC,I-F1,I-AP}
& \metric{I-AUROC,I-F1,I-AP}
& \metric{I-AUROC,I-F1,I-AP}
& \multicolumn{4}{|c@{}}{\avgmetric{I-AUROC,I-F1,I-AP}} \\
\xmark & WinCLIP
& \resimg{90.4}{92.7}{95.6}
& \resimg{75.6}{78.2}{78.8}
& \resimg{68.2}{67.8}{70.9}
& \resimg{93.5}{86.4}{94.2}
& \resimg{91.8}{75.8}{79.5}
& \resimg{95.1}{94.1}{97.7}
& \multicolumn{4}{|c@{}}{\IndAvgImg{85.8}{82.5}{86.1}} \\
\xmark & APRIL-GAN
& \resimg{86.1}{90.4}{93.6}
& \resimg{77.4}{78.6}{80.9}
& \resimg{73.7}{68.7}{69.9}
& \resimg{90.4}{82.9}{92.0}
& \resimg{94.4}{80.3}{83.9}
& \resimg{85.5}{89.1}{94.0}
& \multicolumn{4}{|c@{}}{\IndAvgImg{84.6}{81.7}{85.7}} \\
\xmark & CLIP-AD
& \resimg{74.1}{86.3}{88.1}
& \resimg{66.2}{74.3}{71.4}
& \resimg{66.7}{65.9}{67.3}
& \resimg{81.7}{75.0}{85.5}
& \resimg{62.1}{37.1}{32.3}
& \resimg{75.1}{86.3}{88.0}
& \multicolumn{4}{|c@{}}{\IndAvgImg{71.0}{70.8}{72.1}} \\
\xmark & AnomalyCLIP
& \resimg{91.6}{92.7}{96.2}
& \resimg{81.0}{80.3}{84.4}
& \resimg{88.7}{86.0}{90.6}
& \resimg{91.9}{84.5}{93.4}
& \resimg{98.0}{90.6}{92.4}
& \resimg{93.7}{94.3}{97.4}
& \multicolumn{4}{|c@{}}{\IndAvgImg{90.8}{88.1}{92.4}} \\
\xmark & AdaCLIP
& \resimg{92.0}{92.7}{96.4}
& \resimg{79.7}{79.6}{83.2}
& \resimg{90.0}{87.2}{91.5}
& \resimg{94.9}{90.3}{96.2}
& \resimg{98.3}{91.5}{94.2}
& \resimg{92.1}{92.4}{96.3}
& \multicolumn{4}{|c@{}}{\IndAvgImg{91.2}{89.0}{93.0}} \\
\xmark & FE-CLIP
& \resimg{91.9}{\textendash}{96.5}
& \resimg{84.6}{\textendash}{86.6}
& \resimg{90.3}{\textendash}{90.0}
& \IndResMiss
& \resimg{97.5}{\textendash}{92.3}
& \resimg{98.3}{\textendash}{99.4}
& \multicolumn{4}{|c@{}}{\IndResMiss} \\
\cmark & VisualAD(CLIP)
& \resimg{92.2}{93.2}{96.7}
& \resimg{84.7}{82.5}{87.6}
& \resimg{94.9}{93.9}{97.0}
& \resimg{98.0}{93.9}{98.3}
& \resimg{99.5}{95.0}{97.8}
& \resimg{97.5}{96.6}{99.1}
& \multicolumn{4}{|c@{}}{\IndAvgImg{\secondmean{94.5}}{\secondmean{92.5}}{\secondmean{96.1}}} \\
\cmark & VisualAD(DINOv2)
& \resimg{90.1}{92.4}{94.8}
& \resimg{83.1}{81.4}{86.8}
& \resimg{88.2}{84.7}{89.7}
& \resimg{97.7}{93.1}{98.1}
& \resimg{93.2}{83.9}{86.1}
& \resimg{91.0}{94.4}{97.4}
& \multicolumn{4}{|c@{}}{\IndAvgImg{90.6}{88.3}{92.2}} \\
\rowcolor{TableOursGray}
\cmark & Ours(CLIP)
& \resimg{93.4}{93.3}{96.9}
& \resimg{86.8}{83.6}{89.1}
& \resimg{96.0}{94.9}{98.4}
& \resimg{98.1}{92.9}{98.4}
& \resimg{99.5}{95.8}{97.8}
& \resimg{97.3}{96.0}{98.9}
& \multicolumn{4}{|c@{}}{\IndAvgImg{\bestmean{95.2}}{\bestmean{92.8}}{\bestmean{96.6}}} \\
\rowcolor{TableOursGray}
\cmark & Ours(DINOv2)
& \resimg{92.0}{93.6}{96.1}
& \resimg{82.0}{80.8}{86.4}
& \resimg{85.0}{87.9}{91.4}
& \resimg{97.4}{93.1}{98.0}
& \resimg{98.0}{89.5}{91.4}
& \resimg{93.4}{96.0}{98.1}
& \multicolumn{4}{|c@{}}{\IndAvgImg{91.3}{90.1}{93.6}} \\
\midrule
\rowcolor{TableGroupGray}
\multicolumn{2}{@{}c}{\sectionshift{Pixel-level}}
& \metric{P-AUROC,P-F1,P-AP,P-AUPRO}
& \metric{P-AUROC,P-F1,P-AP,P-AUPRO}
& \metric{P-AUROC,P-F1,P-AP,P-AUPRO}
& \metric{P-AUROC,P-F1,P-AP,P-AUPRO}
& \metric{P-AUROC,P-F1,P-AP,P-AUPRO}
& \metric{P-AUROC,P-F1,P-AP,P-AUPRO}
& \multicolumn{4}{|c@{}}{\avgmetric{P-AUROC,P-F1,P-AP,P-AUPRO}} \\
\xmark & WinCLIP
& \respix{82.3}{24.8}{18.2}{62.0}
& \respix{73.2}{9.0}{5.4}{51.1}
& \respix{72.7}{18.5}{12.9}{27.3}
& \respix{94.1}{24.6}{17.4}{77.6}
& \respix{87.6}{12.7}{6.8}{65.7}
& \respix{79.5}{16.1}{9.8}{51.5}
& \multicolumn{4}{|c@{}}{\IndAvgPix{81.6}{17.6}{11.8}{55.9}} \\
\xmark & APRIL-GAN
& \respix{87.5}{42.3}{39.1}{43.7}
& \respix{93.8}{32.6}{26.2}{86.5}
& \respix{91.3}{40.1}{37.7}{21.0}
& \respix{94.5}{64.2}{66.9}{39.2}
& \respix{84.4}{35.1}{27.8}{12.7}
& \respix{94.9}{60.4}{61.0}{33.8}
& \multicolumn{4}{|c@{}}{\IndAvgPix{91.0}{45.8}{43.1}{39.5}} \\
\xmark & CLIP-AD
& \respix{77.9}{26.3}{21.1}{55.7}
& \respix{93.0}{24.1}{17.9}{80.2}
& \respix{80.9}{24.1}{18.3}{41.4}
& \respix{95.6}{43.0}{39.6}{73.5}
& \respix{69.1}{20.9}{14.7}{36.1}
& \respix{86.6}{35.8}{31.0}{63.2}
& \multicolumn{4}{|c@{}}{\IndAvgPix{83.9}{29.0}{23.8}{58.4}} \\
\xmark & AnomalyCLIP
& \respix{91.0}{38.9}{34.4}{81.7}
& \respix{95.4}{27.6}{20.7}{86.4}
& \respix{93.0}{47.1}{41.5}{71.0}
& \respix{98.0}{50.6}{43.7}{90.8}
& \respix{96.9}{56.9}{53.6}{89.2}
& \respix{97.5}{55.8}{52.5}{87.9}
& \multicolumn{4}{|c@{}}{\IndAvgPix{\secondmean{95.3}}{46.2}{41.1}{84.5}} \\
\xmark & AdaCLIP
& \respix{88.5}{43.9}{41.0}{47.6}
& \respix{95.1}{33.8}{29.2}{71.3}
& \respix{87.7}{42.3}{36.6}{17.1}
& \respix{96.1}{59.2}{58.6}{40.8}
& \respix{88.6}{48.5}{42.6}{37.6}
& \respix{95.1}{58.4}{56.1}{34.3}
& \multicolumn{4}{|c@{}}{\IndAvgPix{91.9}{47.7}{44.0}{41.5}} \\
\xmark & FE-CLIP
& \respix{92.6}{\textendash}{\textendash}{88.3}
& \respix{95.9}{\textendash}{\textendash}{92.8}
& \respix{95.6}{\textendash}{\textendash}{80.4}
& \IndResMiss
& \respix{98.5}{\textendash}{\textendash}{96.6}
& \respix{99.0}{\textendash}{\textendash}{97.4}
& \multicolumn{4}{|c@{}}{\IndResMiss} \\
\cmark & VisualAD(CLIP)
& \respix{90.8}{43.9}{41.2}{87.5}
& \respix{95.8}{34.6}{28.4}{91.0}
& \respix{91.1}{49.8}{43.1}{80.4}
& \respix{98.5}{60.9}{62.1}{98.5}
& \respix{92.2}{57.9}{56.4}{89.3}
& \respix{98.1}{64.3}{65.5}{94.8}
& \multicolumn{4}{|c@{}}{\IndAvgPix{94.4}{51.9}{49.5}{\secondmean{90.3}}} \\
\cmark & VisualAD(DINOv2)
& \respix{91.3}{47.4}{45.4}{88.6}
& \respix{95.3}{35.2}{29.9}{88.2}
& \respix{93.4}{42.6}{38.7}{76.7}
& \respix{98.9}{64.0}{66.8}{98.9}
& \respix{89.5}{54.8}{52.2}{84.5}
& \respix{96.7}{65.8}{67.7}{92.4}
& \multicolumn{4}{|c@{}}{\IndAvgPix{94.2}{51.6}{50.1}{88.2}} \\
\rowcolor{TableOursGray}
\cmark & Ours(CLIP)
& \respix{91.5}{45.0}{43.6}{88.3}
& \respix{95.8}{35.2}{28.5}{90.8}
& \respix{93.8}{51.2}{45.2}{80.9}
& \respix{99.5}{61.3}{63.9}{98.9}
& \respix{95.8}{57.1}{56.4}{92.0}
& \respix{98.1}{62.1}{63.6}{94.4}
& \multicolumn{4}{|c@{}}{\IndAvgPix{\bestmean{95.8}}{\secondmean{52.0}}{\secondmean{50.2}}{\bestmean{90.9}}} \\
\rowcolor{TableOursGray}
\cmark & Ours(DINOv2)
& \respix{91.1}{48.7}{47.5}{88.1}
& \respix{95.9}{36.0}{30.6}{88.0}
& \respix{94.5}{48.8}{45.6}{75.9}
& \respix{99.3}{64.0}{68.0}{98.0}
& \respix{95.7}{60.8}{60.4}{91.8}
& \respix{98.2}{66.9}{70.2}{94.2}
& \multicolumn{4}{|c@{}}{\IndAvgPix{\bestmean{95.8}}{\bestmean{54.2}}{\bestmean{53.7}}{89.3}} \\
\bottomrule
\end{tabularx}
\caption{
Comparisons with state-of-the-art zero-shot AD methods on industrial benchmarks.
\protect\cmark{}/\protect\xmark{} denote language-free/language-based methods.
Image-/pixel-level metrics are
$(\mathrm{AUROC},\mathrm{F1\mbox{-}max},\mathrm{AP})$ /
$(\mathrm{AUROC},\mathrm{F1\mbox{-}max},\mathrm{AP},\mathrm{AUPRO})$.
}
\label{tab:industrial_compact_results}
\endgroup

% ===================== Table 2: Medical =====================
\begingroup
\fontsize{5.8pt}{7.1pt}\selectfont
\setlength{\tabcolsep}{0.85pt}
\renewcommand{\arraystretch}{1.17}
\setlength{\arrayrulewidth}{0.32pt}
\arrayrulecolor{black}
\setlength{\minrowclearance}{0pt}
\setlength{\extrarowheight}{0pt}

\begin{tabularx}{\textwidth}{@{}
>{\centering\arraybackslash}p{0.22cm}
>{\raggedright\arraybackslash}p{1.62cm}
*{4}{>{\hsize=0.93\hsize\centering\arraybackslash}X}
>{\hsize=1.28\hsize\centering\arraybackslash}X
|
*{4}{>{\hsize=0.93\hsize\centering\arraybackslash}X}
>{\hsize=1.28\hsize\centering\arraybackslash}X
@{}}
\toprule
&
\textbf{Method}
& \textbf{OCT17}
& \textbf{BrainMRI}
& \textbf{Brain AD$_I$}
& \textbf{HIS}
& \textbf{I-Avg.}
& \textbf{Brain AD$_P$}
& \textbf{ClinicDB}
& \textbf{Endo}
& \textbf{Kvasir}
& \textbf{P-Avg.} \\
\midrule
\rowcolor{TableGroupGray}
\multicolumn{2}{@{}c}{\textit{\textbf{Image-/Pixel-level}}}
& \metric{I-AUROC,I-AP}
& \metric{I-AUROC,I-AP}
& \metric{I-AUROC,I-AP}
& \metric{I-AUROC,I-AP}
& \avgmetric{I-AUROC,I-F1,I-AP}
& \metric{P-AUROC,P-AP}
& \metric{P-AUROC,P-AP}
& \metric{P-AUROC,P-AP}
& \metric{P-AUROC,P-AP}
& \avgmetric{P-AUROC,P-F1,P-AP,P-AUPRO} \\
\xmark & WinCLIP
& \respair{55.2}{81.1}
& \respair{86.6}{91.5}
& \respair{72.5}{91.6}
& \respair{47.1}{40.9}
& \AvgOneImg{65.4}{75.8}{76.3}
& \respair{87.6}{13.3}
& \respair{70.3}{19.4}
& \respair{68.2}{23.8}
& \respair{69.7}{27.8}
& \AvgOnePix{74.0}{29.5}{21.1}{36.3} \\
\xmark & APRIL-GAN
& \respair{30.4}{67.0}
& \respair{89.3}{90.9}
& \respair{58.8}{87.8}
& \respair{59.8}{56.5}
& \AvgOneImg{59.6}{82.0}{75.6}
& \respair{83.6}{35.0}
& \respair{82.4}{36.4}
& \respair{82.7}{47.2}
& \respair{77.6}{42.5}
& \AvgOnePix{81.6}{41.8}{40.3}{47.0} \\
\xmark & CLIP-AD
& \respair{58.1}{84.1}
& \respair{83.0}{89.3}
& \respair{72.1}{91.5}
& \respair{44.7}{46.0}
& \AvgOneImg{64.5}{80.9}{77.7}
& \respair{94.1}{40.5}
& \respair{76.5}{24.6}
& \respair{78.1}{36.8}
& \respair{73.6}{31.1}
& \AvgOnePix{80.6}{38.8}{33.3}{51.0} \\
\xmark & AnomalyCLIP
& \respair{63.7}{86.5}
& \respair{96.4}{97.2}
& \respair{69.0}{90.1}
& \respair{55.2}{56.1}
& \AvgOneImg{71.1}{84.5}{82.5}
& \respair{95.1}{42.3}
& \respair{84.6}{41.7}
& \respair{86.5}{50.7}
& \respair{82.0}{43.2}
& \AvgOnePix{87.1}{48.1}{44.5}{63.3} \\
\xmark & AdaCLIP
& \respair{77.3}{91.8}
& \respair{94.9}{96.8}
& \respair{80.0}{94.1}
& \respair{59.9}{55.8}
& \AvgOneImg{78.0}{84.5}{84.6}
& \respair{95.2}{37.0}
& \respair{84.3}{43.3}
& \respair{82.9}{47.7}
& \respair{80.3}{31.0}
& \AvgOnePix{85.7}{45.2}{39.8}{45.8} \\
\cmark & VisualAD(CLIP)
& \respair{88.9}{96.4}
& \respair{96.7}{97.6}
& \respair{80.8}{94.7}
& \respair{60.1}{55.9}
& \AvgOneImg{81.6}{\secondmean{85.9}}{86.2}
& \respair{95.2}{43.7}
& \respair{85.2}{37.8}
& \respair{84.9}{46.2}
& \respair{80.3}{47.6}
& \AvgOnePix{86.4}{48.0}{43.8}{66.8} \\
\cmark & VisualAD(DINOv2)
& \respair{91.2}{97.1}
& \respair{93.8}{95.9}
& \respair{87.1}{96.7}
& \respair{60.1}{56.2}
& \AvgOneImg{83.1}{85.5}{86.5}
& \respair{96.4}{51.9}
& \respair{85.9}{36.2}
& \respair{86.8}{53.2}
& \respair{82.6}{46.4}
& \AvgOnePix{\secondmean{87.9}}{50.1}{\secondmean{46.9}}{64.8} \\
\rowcolor{TableOursGray}[\dimexpr\tabcolsep+0.02pt\relax]
\cmark & Ours(CLIP)
& \respair{95.0}{98.4}
& \respair{96.6}{97.7}
& \respair{83.2}{95.6}
& \respair{59.6}{55.5}
& \AvgOneImg{\secondmean{83.6}}{\bestmean{87.1}}{\secondmean{86.8}}
& \respair{95.7}{49.9}
& \respair{86.7}{40.5}
& \respair{86.7}{49.7}
& \respair{82.2}{44.6}
& \AvgOnePix{87.8}{\secondmean{50.4}}{46.2}{\secondmean{67.7}} \\
\rowcolor{TableOursGray}[\dimexpr\tabcolsep+0.02pt\relax]
\cmark & Ours(DINOv2)
& \respair{93.7}{98.1}
& \respair{92.9}{95.6}
& \respair{88.9}{97.2}
& \respair{62.5}{58.1}
& \AvgOneImg{\bestmean{84.5}}{85.5}{\bestmean{87.3}}
& \respair{97.2}{55.4}
& \respair{87.7}{43.3}
& \respair{88.6}{56.6}
& \respair{83.7}{50.0}
& \AvgOnePix{\bestmean{89.3}}{\bestmean{53.2}}{\bestmean{51.3}}{\bestmean{68.6}} \\
\bottomrule
\end{tabularx}
\caption{
Comparisons with state-of-the-art zero-shot AD methods on medical benchmarks.
Classification and segmentation datasets report $(\mathrm{AUROC},\mathrm{AP})$
at image and pixel levels, respectively; mean columns additionally include
$\mathrm{F1\mbox{-}max}$ and $\mathrm{AUPRO}$.
}
\label{tab:medical_compact_results}

\endgroup
\end{table*}

\subsection{Ablation Studies}
\label{sec:ablation_studies}

In this section, we use CLIP’s ViT-L/14@336px backbone by default and conduct ablation studies on the VisA dataset to examine how different settings affect the proposed FreqAnchorAD framework. 
We additionally include DINOv2-based ablations for part of the experiments, with the full results deferred to the appendix.

\subsubsection{Effect of Core Components.}
\label{sec:component_ablation}
Table~\ref{tab:component_ablation_cross_dataset} evaluates the three core components.
Removing FDAP causes the largest degradation in both transfer directions, while LFCM mainly benefits localization and AAS contributes more to image-level discrimination.

\begin{table}[t]
\centering

\scriptsize
\setlength{\tabcolsep}{0.8pt}
\renewcommand{\arraystretch}{1.03}

\begin{tabularx}{\columnwidth}{
@{}
>{\raggedright\arraybackslash}p{0.28\columnwidth}
*{4}{>{\centering\arraybackslash}X}
@{}
}
\toprule
\textbf{Method}
& \textbf{P-AUROC}
& \textbf{P-AP}
& \textbf{I-AUROC}
& \textbf{I-AP} \\
\midrule

\rowcolor{gray!15}
\multicolumn{5}{c}{\textbf{VisA$\rightarrow$MVTec AD}} \\

Full model
& \textbf{91.5}
& \textbf{43.6}
& \textbf{93.4}
& \textbf{96.9} \\

w/o FDAP
& \mbox{88.8{\tiny\color{gray}($\downarrow$2.7)}}
& \mbox{34.8{\tiny\color{gray}($\downarrow$8.8)}}
& \mbox{89.1{\tiny\color{gray}($\downarrow$4.3)}}
& \mbox{94.7{\tiny\color{gray}($\downarrow$2.2)}} \\

w/o LFCM
& \mbox{\underline{91.1}{\tiny\color{gray}($\downarrow$0.4)}}
& \mbox{42.6{\tiny\color{gray}($\downarrow$1.0)}}
& \mbox{\underline{92.4}{\tiny\color{gray}($\downarrow$1.0)}}
& \mbox{\underline{96.7}{\tiny\color{gray}($\downarrow$0.2)}} \\

w/o AAS
& \mbox{\underline{91.1}{\tiny\color{gray}($\downarrow$0.4)}}
& \mbox{\underline{42.8}{\tiny\color{gray}($\downarrow$0.8)}}
& \mbox{91.6{\tiny\color{gray}($\downarrow$1.8)}}
& \mbox{96.3{\tiny\color{gray}($\downarrow$0.6)}} \\

\midrule

\rowcolor{gray!15}
\multicolumn{5}{c}{\textbf{MVTec AD$\rightarrow$VisA}} \\

Full model
& \textbf{95.8}
& \textbf{28.5}
& \underline{86.8}
& \textbf{89.1} \\

w/o FDAP
& \mbox{94.2{\tiny\color{gray}($\downarrow$1.6)}}
& \mbox{23.3{\tiny\color{gray}($\downarrow$5.2)}}
& \mbox{85.5{\tiny\color{gray}($\downarrow$1.3)}}
& \mbox{87.2{\tiny\color{gray}($\downarrow$1.9)}} \\

w/o LFCM
& \mbox{95.4{\tiny\color{gray}($\downarrow$0.4)}}
& \mbox{26.7{\tiny\color{gray}($\downarrow$1.8)}}
& \mbox{\textbf{87.2}{\tiny\color{gray}($\uparrow$0.4)}}
& \mbox{\underline{88.9}{\tiny\color{gray}($\downarrow$0.2)}} \\

w/o AAS
& \mbox{\underline{95.7}{\tiny\color{gray}($\downarrow$0.1)}}
& \mbox{\underline{28.3}{\tiny\color{gray}($\downarrow$0.2)}}
& \mbox{86.4{\tiny\color{gray}($\downarrow$0.4)}}
& \mbox{88.7{\tiny\color{gray}($\downarrow$0.4)}} \\

\bottomrule
\end{tabularx}

\caption{
Ablation study of LFCM, FDAP, and AAS for cross-dataset transfer between MVTec AD and VisA.
Parentheses report changes relative to the full model.
}
\label{tab:component_ablation_cross_dataset}
\end{table}

\subsubsection{Effect of Canonicalization Ratio.}
\label{sec:canonicalization_analysis}
We analyze the training-time soft canonicalization coefficient $\rho$, which interpolates between the pretrained channel order and the source-derived variance order.
Table~\ref{tab:perm_ablation} shows stable performance across a broad range, indicating low sensitivity to $\rho$.
We use $\rho=0.6$ during training and $\rho=1.0$ at inference to align samples with the source-derived channel coordinate.

\begin{table}[t]
\centering

\tiny
\setlength{\tabcolsep}{0.8pt}
\renewcommand{\arraystretch}{0.96}
\setlength{\aboverulesep}{0pt}
\setlength{\belowrulesep}{0pt}

% Gray color for selected row
\definecolor{OursRowGray}{RGB}{230,230,230}

\newcommand{\PermGroupCellTightFixed}[3]{%
  \begingroup
  \renewcommand{\arraystretch}{0.78}%
  \begin{tabular}[c]{@{}c@{}}
    \rule{0pt}{2.25ex}%
    \textbf{\textit{#1}}\\[-0.28ex]
    \rule{2.94cm}{0.30pt}\\[-0.32ex]
    \makebox[2.94cm][c]{%
      \makebox[1.47cm][c]{\textbf{\textit{#2}}}%
      \makebox[1.47cm][c]{\textbf{\textit{#3}}}%
    }%
    \rule[-0.55ex]{0pt}{0pt}%
  \end{tabular}%
  \endgroup
}

\begin{adjustbox}{width=\columnwidth}
\begin{tabular}{
@{}
>{\centering\arraybackslash}p{0.45cm}|
*{4}{>{\centering\arraybackslash}p{1.47cm}}
@{}
}
\toprule[0.9pt]

\textbf{$\rho$}
& \multicolumn{2}{c}{
    \PermGroupCellTightFixed{VisA $\rightarrow$ MVTec AD}{Pixel}{Image}
  }
& \multicolumn{2}{c}{
    \PermGroupCellTightFixed{MVTec AD $\rightarrow$ VisA}{Pixel}{Image}
  } \\
\midrule[0.45pt]

0.2
& (91.2,43.3,39.3)
& (\underline{93.5},92.8,96.5)
& (94.9,30.7,24.2)
& (85.3,82.5,88.1) \\

0.4
& (\textbf{91.5},\textbf{45.0},\underline{43.1})
& (\textbf{93.8},\textbf{93.3},\textbf{97.0})
& (\underline{95.6},\underline{34.7},28.0)
& (\underline{86.7},\textbf{83.7},\textbf{89.1}) \\

\rowcolor{OursRowGray}
\textbf{0.6}
& (\textbf{91.5},\textbf{45.0},\textbf{43.6})
& (93.4,\textbf{93.3},\underline{96.9})
& (\textbf{95.8},\textbf{35.2},\textbf{28.5})
& (\textbf{86.8},\underline{83.6},\textbf{89.1}) \\

0.8
& (\underline{91.4},\underline{44.9},\textbf{43.6})
& (93.1,\underline{93.2},96.8)
& (\textbf{95.8},\underline{34.7},\underline{28.2})
& (86.6,83.4,\underline{88.9}) \\

1.0
& (\underline{91.4},44.8,\textbf{43.6})
& (93.0,93.1,96.8)
& (\textbf{95.8},34.4,27.9)
& (86.5,83.2,\underline{88.9}) \\

\bottomrule[0.9pt]
\end{tabular}
\end{adjustbox}
\caption{
Ablation on the training-time soft canonicalization coefficient $\rho$.
All results use $\rho=1$ during inference.
For each transfer direction, pixel-level and image-level metrics are reported as (AUROC, F1-max, AP).
}
\label{tab:perm_ablation}
\end{table}

\subsubsection{Effect of Frequency-Deviation Anchor Projection.}
\label{sec:fdap_analysis}
Table~\ref{tab:lfcm_fdap_ablation} shows that LFCM substantially improves the raw token space, while a plain MLP provides further gains.
Compared with the plain MLP, FDAP improves seven of eight metrics and matches the remaining one, demonstrating consistent benefits beyond generic nonlinear projection.

\begin{table}[t]
\centering

\scriptsize
\setlength{\tabcolsep}{0.8pt}
\renewcommand{\arraystretch}{1.05}

\begin{tabularx}{\columnwidth}{
@{}
>{\raggedright\arraybackslash}p{0.28\columnwidth}
*{4}{>{\centering\arraybackslash}X}
@{}
}
\toprule
\textbf{Space}
& \textbf{P-AUROC}
& \textbf{P-AP}
& \textbf{I-AUROC}
& \textbf{I-AP} \\
\midrule

\rowcolor{gray!15}
\multicolumn{5}{c}{\textbf{VisA$\rightarrow$MVTec AD}} \\

LFCM + FDAP
& \textbf{91.1}
& \textbf{42.8}
& \textbf{91.6}
& \textbf{96.3} \\

LFCM + Plain MLP
& \mbox{\underline{90.9}{\tiny\color{gray}($\downarrow$0.2)}}
& \mbox{\underline{42.6}{\tiny\color{gray}($\downarrow$0.2)}}
& \mbox{\underline{91.4}{\tiny\color{gray}($\downarrow$0.2)}}
& \mbox{\underline{96.1}{\tiny\color{gray}($\downarrow$0.2)}} \\

LFCM token space
& \mbox{89.7{\tiny\color{gray}($\downarrow$1.4)}}
& \mbox{38.2{\tiny\color{gray}($\downarrow$4.6)}}
& \mbox{90.8{\tiny\color{gray}($\downarrow$0.8)}}
& \mbox{95.8{\tiny\color{gray}($\downarrow$0.5)}} \\

Raw token space
& \mbox{83.6{\tiny\color{gray}($\downarrow$7.5)}}
& \mbox{20.6{\tiny\color{gray}($\downarrow$22.2)}}
& \mbox{77.9{\tiny\color{gray}($\downarrow$13.7)}}
& \mbox{88.9{\tiny\color{gray}($\downarrow$7.4)}} \\

\midrule

\rowcolor{gray!15}
\multicolumn{5}{c}{\textbf{MVTec AD$\rightarrow$VisA}} \\

LFCM + FDAP
& \textbf{95.7}
& \textbf{28.3}
& \underline{86.4}
& \underline{88.7} \\

LFCM + Plain MLP
& \mbox{\textbf{95.7}{\tiny\color{gray}($\downarrow$0.0)}}
& \mbox{\underline{27.5}{\tiny\color{gray}($\downarrow$0.8)}}
& \mbox{85.6{\tiny\color{gray}($\downarrow$0.8)}}
& \mbox{88.3{\tiny\color{gray}($\downarrow$0.4)}} \\

LFCM token space
& \mbox{\underline{94.5}{\tiny\color{gray}($\downarrow$1.2)}}
& \mbox{23.8{\tiny\color{gray}($\downarrow$4.5)}}
& \mbox{85.3{\tiny\color{gray}($\downarrow$1.1)}}
& \mbox{86.9{\tiny\color{gray}($\downarrow$1.8)}} \\

Raw token space
& \mbox{94.0{\tiny\color{gray}($\downarrow$1.7)}}
& \mbox{22.3{\tiny\color{gray}($\downarrow$6.0)}}
& \mbox{\textbf{87.5}{\tiny\color{gray}($\uparrow$1.1)}}
& \mbox{\textbf{89.4}{\tiny\color{gray}($\uparrow$0.7)}} \\

\bottomrule
\end{tabularx}

\caption{
Bidirectional cross-dataset ablation of LFCM and FDAP, with AAS disabled.
Parentheses show changes relative to LFCM + FDAP in each direction.
}
\label{tab:lfcm_fdap_ablation}
\end{table}

\subsubsection{Progressive Separation Across Representation Spaces.}
\label{sec:fdap_separation}
\begin{figure}[t]
\centering
\includegraphics[width=\columnwidth]{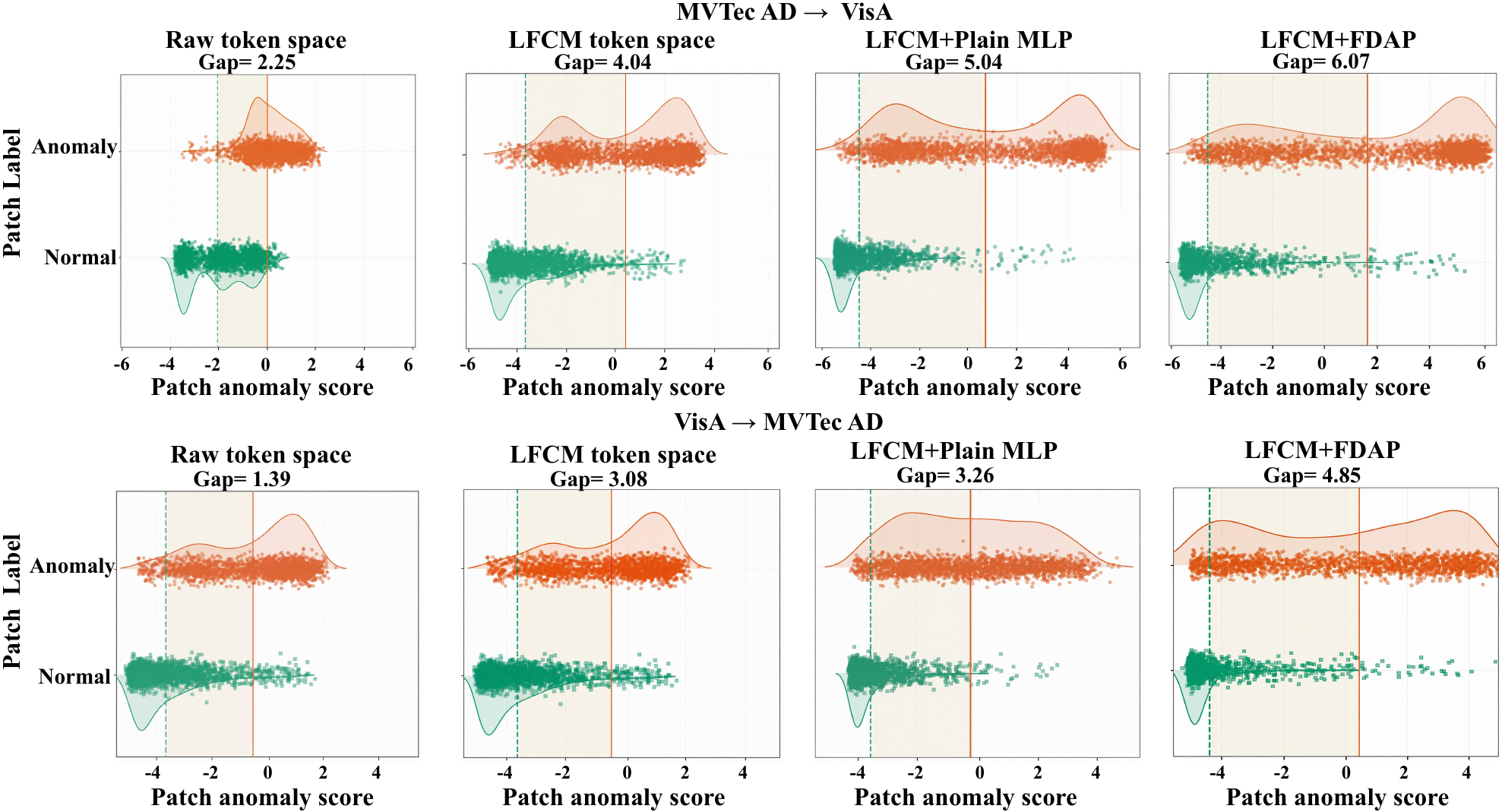}
\caption{
Patch-score distributions across representation spaces in both directions, with AAS disabled.
Columns show raw tokens, LFCM, plain MLP, and FDAP; green and orange denote normal and anomalous patches, with vertical lines marking their means and shading indicating the gap.
}
\label{fig:fdap_score_separation}
\label{fig:lfcm+fdap_visual}
\end{figure}

Figure~\ref{fig:fdap_score_separation} shows progressively improved separation between normal and anomalous patches across representation spaces.
FDAP achieves the largest mean-score gaps in both directions, showing an advantage over a plain MLP.

\subsubsection{Effect of Spectral Layout.}
% \label{sec:spectral_layout}
Table~\ref{tab:dct_layout} shows that the source-derived channel order remains effective under both 1D and 2D organization.
The anisotropic $16\times64$ layout performs worst, whereas the balanced $32\times32$ layout provides the best overall result and is therefore adopted by default.

\begin{table}[t]
\centering

\scriptsize
\setlength{\tabcolsep}{1.2pt}
\renewcommand{\arraystretch}{0.96}
\setlength{\doublerulesep}{1.1pt}

\setlength{\aboverulesep}{0pt}
\setlength{\belowrulesep}{0pt}

\definecolor{LayoutGray}{RGB}{215,215,215}

\begin{tabularx}{\columnwidth}{
@{}
>{\centering\arraybackslash}p{0.36\columnwidth}||
*{4}{>{\centering\arraybackslash}X}
@{}
}
\toprule[0.9pt]

\textbf{\textit{Spectral Organization}}
& \textbf{\textit{P-AUROC}}
& \textbf{\textit{P-AP}}
& \textbf{\textit{I-AUROC}}
& \textbf{\textit{I-AP}} \\
\midrule[0.45pt]

\rowcolor{LayoutGray}
\textbf{2D grid ($32\times32$)}
& \textbf{91.5}
& \textbf{43.6}
& \textbf{93.4}
& \underline{96.9} \\

\textbf{1D sequence}
& \mbox{\underline{91.1}{\tiny\color{gray}($\downarrow$0.4)}}
& \mbox{42.6{\tiny\color{gray}($\downarrow$1.0)}}
& \mbox{93.0{\tiny\color{gray}($\downarrow$0.4)}}
& \mbox{96.7{\tiny\color{gray}($\downarrow$0.2)}} \\

\textbf{2D grid ($16\times64$)}
& \mbox{89.6{\tiny\color{gray}($\downarrow$1.9)}}
& \mbox{40.1{\tiny\color{gray}($\downarrow$3.5)}}
& \mbox{91.1{\tiny\color{gray}($\downarrow$2.3)}}
& \mbox{96.1{\tiny\color{gray}($\downarrow$0.8)}} \\

\textbf{2D grid ($64\times16$)}
& \mbox{\underline{91.1}{\tiny\color{gray}($\downarrow$0.4)}}
& \mbox{\underline{42.8}{\tiny\color{gray}($\downarrow$0.8)}}
& \mbox{\underline{93.3}{\tiny\color{gray}($\downarrow$0.1)}}
& \mbox{\textbf{97.0}{\tiny\color{gray}($\uparrow$0.1)}} \\

\bottomrule[0.9pt]
\end{tabularx}
\caption{
Ablation of spectral organization in FDAP under VisA$\rightarrow$MVTec AD cross-dataset evaluation.
Parentheses show changes from the default 2D grid ($32\times32$).
}
\label{tab:dct_layout}
\end{table}

\subsubsection{Effect of Transform Basis.}
Table~\ref{tab:transform_basis} shows that 2D DCT provides the best overall trade-off across pixel- and image-level metrics.
It leads or ties the image-level results and remains within 0.1 points of the best pixel-level scores.
Although Identity and Hadamard remain competitive on individual metrics, PCA and random orthogonal transforms produce larger drops, supporting the ordered cosine basis of DCT as a stable and interpretable default.

\begin{table}[t]
\centering

\scriptsize
\setlength{\tabcolsep}{1.2pt}
\renewcommand{\arraystretch}{0.96}
\setlength{\doublerulesep}{1.1pt}

\setlength{\aboverulesep}{0pt}
\setlength{\belowrulesep}{0pt}

% Gray color for the selected row
\definecolor{BasisRowGray}{RGB}{215,215,215}

\begin{tabularx}{\columnwidth}{
@{}
>{\centering\arraybackslash}p{0.36\columnwidth}||
*{4}{>{\centering\arraybackslash}X}
@{}
}
\toprule[0.9pt]

\textbf{\textit{Transform Basis}}
& \textbf{\textit{P-AUROC}}
& \textbf{\textit{P-AP}}
& \textbf{\textit{I-AUROC}}
& \textbf{\textit{I-AP}} \\
\midrule[0.45pt]

\rowcolor{BasisRowGray}
\textbf{2D DCT ($32\times32$)}
& \underline{91.5}
& \underline{43.6}
& \textbf{93.4}
& \textbf{96.9} \\

\textbf{Normalized Hadamard}
& \mbox{91.4{\tiny\color{gray}($\downarrow$0.1)}}
& \mbox{\textbf{43.7}{\tiny\color{gray}($\uparrow$0.1)}}
& \mbox{\underline{93.3}{\tiny\color{gray}($\downarrow$0.1)}}
& \mbox{\textbf{96.9}{\tiny\color{gray}($\downarrow$0.0)}} \\

\textbf{Identity}
& \mbox{\textbf{91.6}{\tiny\color{gray}($\uparrow$0.1)}}
& \mbox{42.9{\tiny\color{gray}($\downarrow$0.7)}}
& \mbox{\underline{93.3}{\tiny\color{gray}($\downarrow$0.1)}}
& \mbox{\textbf{96.9}{\tiny\color{gray}($\downarrow$0.0)}} \\

\textbf{PCA rotation}
& \mbox{89.9{\tiny\color{gray}($\downarrow$1.6)}}
& \mbox{40.6{\tiny\color{gray}($\downarrow$3.0)}}
& \mbox{92.7{\tiny\color{gray}($\downarrow$0.7)}}
& \mbox{96.5{\tiny\color{gray}($\downarrow$0.4)}} \\

\textbf{Random orthogonal}
& \mbox{91.3{\tiny\color{gray}($\downarrow$0.2)}}
& \mbox{42.9{\tiny\color{gray}($\downarrow$0.7)}}
& \mbox{92.9{\tiny\color{gray}($\downarrow$0.5)}}
& \mbox{\underline{96.7}{\tiny\color{gray}($\downarrow$0.2)}} \\

\bottomrule[0.9pt]
\end{tabularx}

\caption{
Transform-basis ablation for FDAP under VisA$\rightarrow$MVTec AD cross-dataset evaluation.
Parentheses show changes from the default 2D DCT ($32\times32$) basis.
}
\label{tab:transform_basis}

\end{table}

\subsection{Visualization of Channel Spectral Bands}
\label{sec:qualitative_analysis}

\begin{figure}[H]
\centering
\includegraphics[width=\columnwidth]{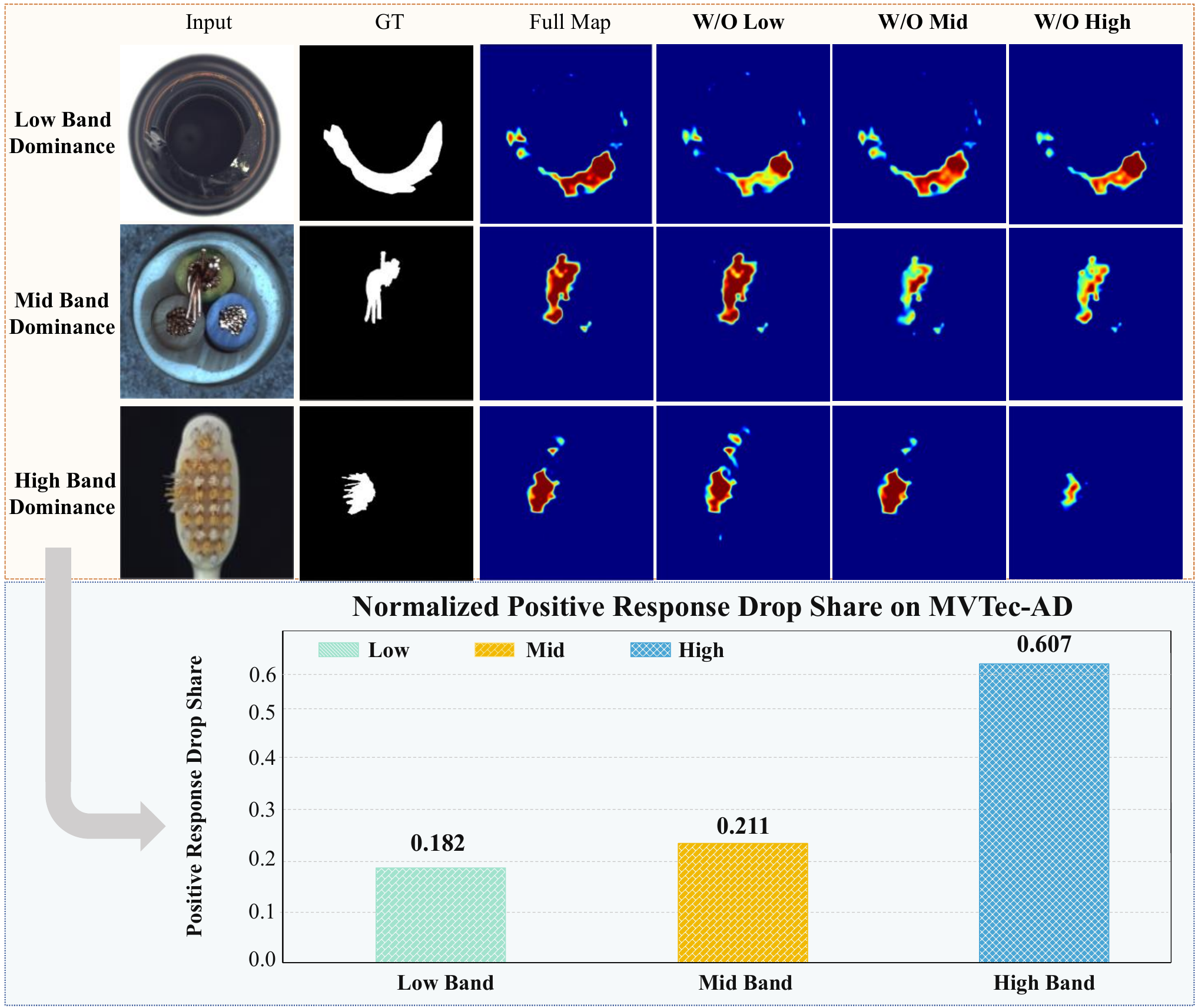}
\caption{
Individual band removal analysis on MVTec AD, visualizing the dominance of different channel spectral bands across images and their normalized positive response drop shares within ground-truth anomaly regions.
}
\label{fig:channel_spectral_orders}
\end{figure}

As shown in Fig.~\ref{fig:channel_spectral_orders}, no single spectral order consistently dominates across samples. Depending on the anomaly appearance, low-, mid-, or high-order responses can each provide the principal localization evidence. Their non-negligible aggregate contributions further demonstrate that the three spectral orders are complementary. Jointly modeling these band-dependent responses therefore preserves more complete anomaly evidence than removing any individual band, validating the effectiveness of multi-band collaboration.

\section{Conclusion}
\label{sec:conclusion}
This paper presents FreqAnchorAD, a language-free frequency-aware framework for zero-shot anomaly detection.
It combines LFCM for local frequency enhancement, FDAP as the core module for anchor projection in a source-derived channel space, and AAS for stable anchor discrimination.
Without text encoders, cross-modal alignment, target-domain training, or test-time adaptation, the framework directly generalizes from auxiliary source categories to unseen industrial and medical categories.
Experiments on thirteen benchmarks demonstrate state-of-the-art mean performance across image-level anomaly recognition and pixel-level defect localization.

\bibliography{main}
\end{document}